\documentclass[conference]{IEEEtran}
\IEEEoverridecommandlockouts

\usepackage{cite}
\usepackage{amsmath,amssymb,amsfonts}
\usepackage{bm}
\usepackage{url}
\usepackage{algorithmic}
\usepackage{graphicx}
\usepackage{textcomp}
\usepackage{xcolor}
\usepackage{multirow}
\usepackage{multicol}
\usepackage{makecell}
\def\BibTeX{{\rm B\kern-.05em{\sc i\kern-.025em b}\kern-.08em
    T\kern-.1667em\lower.7ex\hbox{E}\kern-.125emX}}
\begin{document}

\title{Acquiring Pronunciation Knowledge from Transcribed Speech Audio via Multi-task Learning
\thanks{This work was supported in part by the UKRI Centre for Doctoral Training in Natural Language Processing, funded by the UKRI (grant EP/S022481/1), the University of Edinburgh, School of Informatics and School of Philosophy, Psychology \& Language Sciences and Huawei.}
}

\author{\IEEEauthorblockN{Siqi Sun}
\IEEEauthorblockA{\textit{The University of Edinburgh} \\
Edinburgh, UK \\
Siqi.Sun@ed.ac.uk}
\and
\IEEEauthorblockN{Korin Richmond}
\IEEEauthorblockA{\textit{The University of Edinburgh} \\
Edinburgh, UK \\
Korin.Richmond@ed.ac.uk}
}

\maketitle

\begin{abstract}
Recent work has shown the feasibility and benefit of bootstrapping an integrated sequence-to-sequence (Seq2Seq) linguistic frontend from a traditional pipeline-based frontend for text-to-speech (TTS). To overcome the fixed lexical coverage of bootstrapping training data, previous work has proposed to leverage easily accessible transcribed speech audio as an additional training source for acquiring novel pronunciation knowledge for uncovered words, which relies on an auxiliary ASR model as part of a cumbersome implementation flow. In this work, we propose an alternative method to leverage transcribed speech audio as an additional training source, based on multi-task learning (MTL). Experiments show that, compared to a baseline Seq2Seq frontend, the proposed MTL-based method reduces PER from 2.5\% to 1.6\% for those word types covered exclusively in transcribed speech audio, achieving a similar performance to the previous method but with a much simpler implementation flow.
\end{abstract}

\begin{IEEEkeywords}
pronunciation learning, knowledge transfer, multi-task learning, linguistic frontend, text-to-speech synthesis.
\end{IEEEkeywords}

\section{Introduction}
\label{introduction}


To ensure pronunciation accuracy, recent text-to-speech (TTS) takes as input pronunciation sequences generated by a separate pipeline-based linguistic frontend that includes a dictionary for word pronunciation lookup \cite{Fong2019, Shen2020, Ren2021, Tan2021}.
More recent work shows the benefit of replacing the pipeline with a unified sequence-to-sequence (Seq2Seq) model that directly converts the \emph{text sequence} (a string of characters) to a \emph{pronunciation sequence} (a string of pronunciation tokens including phones, lexical stresses, prosodic boundaries, etc.) at the sentence level (e.g., converting \emph{PIPER'S SON} to \emph{1 p ai p - 0 @ z + 1 s uh n \_B} in Unisyn \cite{Fitt2000} tokens) \cite{Conkie2020, Pan2020, Sun2023, Comini2023}. Due to the lack of ground-truth training targets, a bootstrapping approach is often applied, where a pre-existing pipeline-based frontend is utilized to generate pronunciation sequences for abundant unlabelled text to serve as training targets. The text should cover a wide range of in-dictionary word types\footnote{We follow the standard definition that a \emph{word token} is a single occurrence of a distinct \emph{word type} in the text.} but omit out-of-dictionary ones to ensure target accuracy\cite{Sun2023}.

However, the dictionary size is fixed, so the bootstrapping training data has fixed lexical coverage, which in turn limits the performance of bootstrapped Seq2Seq frontend \cite{Sun2023}. To solve this, we can turn to some additional training source to acquire pronunciation knowledge of certain word types that are not covered in the bootstrapping training data, where the knowledge can be encoded in some form other than pronunciation sequences.
E.g., a Forced-Alignment (FA) method was proposed in \cite{Sun2023} to leverage transcribed speech audio (i.e., pairs of text and speech audio) as an additional training source.
Though effective, the method requires training specific automatic speech recognition (ASR) models as part of a cumbersome pre-train$\rightarrow$ASR-train\&decode$\rightarrow$re-train flow.

In this work, we propose an alternative method to leverage transcribed speech audio as an additional training source, via multi-task learning (MTL).
MTL utilizes training targets of related \emph{extra tasks} as an inductive bias to improve the generalization on the \emph{main task}, by jointly learning the main task and extra tasks using a shared representation \cite{Caruana1998}.
Recently, \cite{Sun2024} further showed in multi-accent frontend modelling, MTL particularly benefits generalizing the \emph{main task} to \emph{extra-exclusive} word types (i.e., word types covered in \emph{extra task} training data but not covered in \emph{main task} training data).
Inspired by this, we propose an MTL-based method, jointly learning the \emph{main task} of Seq2Seq frontend modelling (trained with bootstrapping data) and the \emph{extra task} of acoustic feature regression (trained with transcribed speech audio).
Our goal is to similarly greatly benefit generalizing frontend modelling to those word types covered exclusively in transcribed speech audio, which has an equivalent effect in \textbf{acquiring pronunciation knowledge from transcribed speech audio for the Seq2Seq frontend}.

Our method has a compact pre-train$\rightarrow$re-train flow, completely avoiding ASR training and decoding.
The contributions of this paper are as follows: 1) We propose a novel MTL-based method to acquire pronunciation knowledge from transcribed speech audio. 2) We propose a novel multi-task model for our method. 3) Our experiments and analyses confirm the effectiveness of this method and model.

\section{Background and Related Work}
\label{background}

\subsection{Seq2Seq Frontend}
\label{seq2seq}

Seq2Seq frontends \cite{Conkie2020, Pan2020, Sun2023, Comini2023, Rezackova2021, Ploujnikov2022} convert the text sequence $\bm{x}_{1:L} = [x_1, \dots, x_L]$ (a string of characters) to the pronunciation sequence $\bm{p}_{1:T} = [p_1, \dots, p_T]$ (a string of pronunciation tokens) at the sentence level , where $L$ and $T$ denote the sequence lengths, respectively.
In an RNN-based implementation \cite{Sun2023}, $\bm{x}_{1:L}$ is encoded by a bi-directional LSTM \textbf{Text Encoder} into the encoding vectors $\bm{e}_{1:L}$.
\begin{equation}
\bm{e}_{1:L} = \text{BiLSTM} (\bm{x}_{1:L})
\end{equation}

The \textbf{Pronunciation Decoder} then converts $\bm{e}_{1:L}$ to $\bm{p}_{1:T}$ through a series of transformations. Concretely, for $t \in [1, T]$,
\begin{align}
\bm{a}_t &= \text{LSTM}_{Att} (\bm{a}_{t-1}, \bm{c}_{t-1}, p_{t-1}) \\
\bm{c}_t &= \text{Attention} (\bm{e}_{1:L}, \bm{a}_t) \label{equ:attn} \\
\bm{d}_t &= \text{LSTM}_{Dec} (\bm{d}_{t-1}, \bm{c}_{t}, \bm{a}_{t}) \\
\bm{l}_t &= \text{Projection} (\bm{d}_t), \hspace{4pt} p_t \sim \text{Softmax} (\bm{l}_t)
\end{align}
where the attention hidden states $\bm{a}_{1:T}$, the context vectors $\bm{c}_{1:T}$, the decoder hidden states $\bm{d}_{1:T}$ and the logit vectors $\bm{l}_{1:T}$ are of the same length as $\bm{p}_{1:T}$. In this work, we adopt monotonic GMM attention (V2) \cite{Battenberg2020} for (\ref{equ:attn}).
As in \cite{Sun2023}, text normalisation (TN) is not included in our Seq2Seq frontend modelling to avoid the interference caused by non-standard words (e.g., abbreviations, numbers, etc.) during MTL.

\subsection{Acquiring Pronunciation from Transcribed Speech Audio}
\label{speech}

Previous work has proposed to acquire pronunciation knowledge from transcribed speech audio to improve various pronunciation models, mostly relying on auxiliary ASR systems to first decode pronunciations from speech audio and then improve the model with transcription-pronunciation pairs \cite{Rutherford2014, Kou2015, Bruguier2016, Bruguier2017, Ribeiro2023}. 
Recently, this approach has been adopted (i.e., the FA method \cite{Sun2023}) to improve a Seq2Seq frontend that is \emph{pre-trained} with bootstrapping training data. With a transcribed speech dataset, the pre-trained Seq2Seq frontend generates for each text sequence i) the 1-best pronunciation sequence for training HMM/GMM-based ASR models, and ii) a list of n-best pronunciation sequences for building a language model. Then, the language model is force-aligned with the corresponding speech audio (i.e., the MFCC feature sequence) by the trained ASR models to decode the closest pronunciation sequence. Finally, the $\langle$text, decoded pronunciation$\rangle$ pairs are used to improve the pre-trained Seq2Seq frontend.

\subsection{Multi-task Learning in Pronunciation Modelling}
\label{mtl}

MTL \cite{Milde2017, Route2019, Thompson2022, Ying2024, Kang2024}, including its specific case of multi-lingual/multi-accent modelling \cite{Peters2017, Route2019, Kim2022, Zhu2022, Conkie2020, Comini2023, Sun2024}, has been used to improve various pronunciation modelling tasks. MTL improves Arabic diacritization by forcing a shared model to both diacritize and translate \cite{Thompson2022}, and improves G2P conversion by modelling multiple languages or multiple phonetic alphabets jointly \cite{Milde2017}. 
Generally, \emph{main task} training data and \emph{extra task} training data do not completely overlap in lexical coverage, and previous studies do not differentiate between the generalization on the \emph{main task} to \emph{extra-exclusive} word types and that to \emph{out-of-vocabulary} (OOV) word types (i.e., ones not covered in any training set during MTL) during evaluation.
Recently, \cite{Sun2024} showed MTL particularly benefits generalizing the \emph{main task} to \emph{extra-exclusive} word types.
Inspired by this, we leverage MTL as a method to transfer pronunciation knowledge of \emph{extra-exclusive} word types contained in transcribed speech audio to the Seq2Seq frontend.

\section{Method}
\label{method}

\begin{figure}[!t]
\centering
\includegraphics[width=0.48\textwidth]{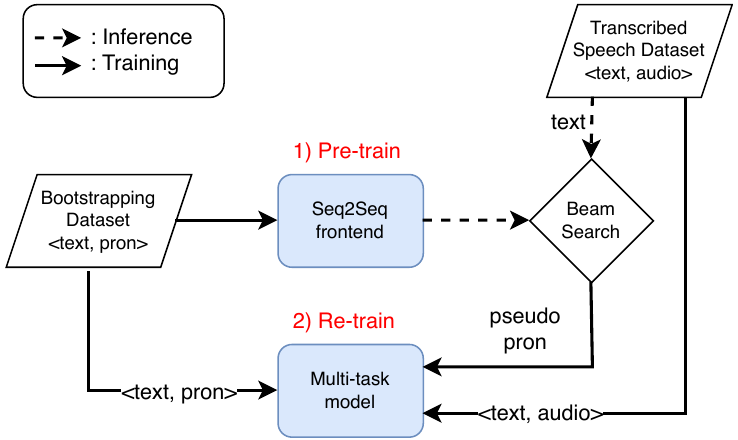}
\caption{The two stages involved in our MTL-based method.}
\label{fig:flowchart}
\end{figure}

During MTL, we set the \emph{main task} to be Seq2Seq frontend modelling and the \emph{extra task} to be the regression on acoustic feature sequences derived from speech audio.
Accordingly, two training sets are involved, including the original \emph{bootstrapping dataset} $\mathcal{D}_{BS} = \{(\bm{x}^{(i)}, \bm{p}^{(i)})\}_{i=1}^N$ for the \emph{main task}, and the \emph{transcribed speech dataset} $\mathcal{D}_{TS} = \{(\bm{x}^{(j)}, \bm{m}^{(j)})\}_{j=1}^M$ for the \emph{extra task}, containing word types not covered by $\mathcal{D}_{BS}$.
Notations $\bm{x}$, $\bm{p}$ and $\bm{m}$ denote the text, pronunciation and acoustic feature sequences respectively, and $N$ and $M$ denote the dataset sizes. 
As shown in Sec.\ \ref{architecture}, $\bm{p}^{(j)}$ corresponding to $\forall \bm{x}^{(j)} \in \mathcal{D}_{TS}$ are required by some MTL setting as an additional input for \emph{extra task}, but are usually inaccessible. To that end, a pre-trained bootstrapped Seq2Seq frontend is leveraged to generate pseudo pronunciation sequences $\bar{\bm{p}}^{(j)}$ for $\forall \bm{x}^{(j)} \in \mathcal{D}_{TS}$ with Beam Search, and then form the pseudo augmented \emph{TS dataset} $\bar{\mathcal{D}}_{TS} = \{(\bm{x}^{(j)}, \bar{\bm{p}}^{(j)}, \bm{m}^{(j)})\}_{j=1}^M$. Therefore, our MTL-based method has a pre-train$\rightarrow$re-train flow, as shown in Fig.\ \ref{fig:flowchart}.

A similar MTL-based method has been proposed to improve a word-level G2P model in \cite{Route2019}. However, they use a parallel dataset during MTL (i.e., tuples of $(\bm{x}, \bm{p}, \bm{m})$, with each tuple corresponding to the same word), so no word type is covered exclusively in transcribed speech audio. Hence, their setting and study aim are totally different from ours. Moreover, our method focuses on the sentence level, which is much harder.

\section{Multi-task Model architecture}
\label{architecture}

\begin{figure}[!t]
\centering
\includegraphics[width=0.48\textwidth]{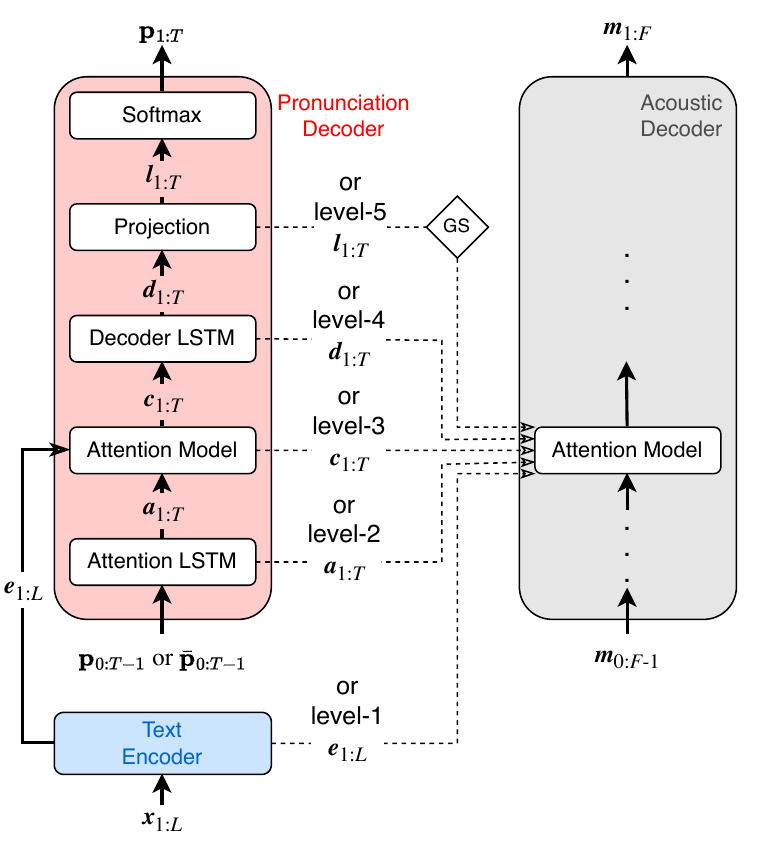}
\caption{The multi-task model architecture. The dashed arrows indicate the Acoustic Decoder is integrated by attending to \emph{only one} of the intermediate representations of the standard Seq2Seq frontend (i.e., Text Encoder + Pronunciation Decoder. See Sec.\ \ref{seq2seq}). GS stands for Gumbel-Softmax.}
\label{fig:multitask_model}
\end{figure}

To implement MTL, we adapt the Seq2Seq frontend (Sec.\ \ref{seq2seq}) by integrating an \textbf{Acoustic Decoder}. The resulting multi-task model is shown in Fig.\ \ref{fig:multitask_model}. The acoustic decoder has a similar model architecture as Tacotron2's decoder plus postnet \cite{Shen2018}, and is equipped with a different attention model. It is tasked with regressing the acoustic feature sequence $\bm{m}_{1:F}$, where $F$ denotes the length in frames.
The MTL loss is:
\begin{align}
\mathcal{L}_{MTL} &= \mathcal{L}_{p} + \lambda * \mathcal{L}_{a} \\
\mathcal{L}_{p} = \mathop{\mbox{$\mathbb{E}$}}_{(\bm{x}, \bm{p})\in \mathcal{D}_{BS}} &\big[-\log P(\bm{p}|\bm{x} ; \bm{\theta}_{e}, \bm{\theta}_{p}) \big] \\
\mathcal{L}_{a} = \mathop{\mbox{$\mathbb{E}$}}_{(\bm{x}, \bar{\bm{p}}, \bm{m})\in \bar{\mathcal{D}}_{TS}} &\big[-\log P(\bm{m}|\bm{x}, \bar{\bm{p}} ; \bm{\theta}_{e}, \hat{\bm{\theta}}_{p}, \bm{\theta}_{a}) \big] 
\end{align}
where $\mathcal{L}_{p}$ is the pronunciation loss averaged over $\mathcal{D}_{BS}$, $\mathcal{L}_{a}$ is the acoustic loss averaged over $\bar{\mathcal{D}}_{TS}$ and  $\lambda$ is a weighting factor. $\mathcal{L}_{p}$ can be reduced to a cross-entropy loss. $\mathcal{L}_{a}$ can be reduced to an L1/L2 loss or remain the negative log-likelihood (NLL) loss when fitting a Laplacian mixture model (LMM) to the underlying distribution \cite{Ren2022, Kogel2023}. 
Minimizing $\mathcal{L}_{p}$ optimizes the encoder $\bm{\theta}_{e}$ and the pronunciation decoder $\bm{\theta}_{p}$, whereas minimizing $\mathcal{L}_{a}$ optimizes $\bm{\theta}_{e}$, the acoustic decoder $\bm{\theta}_{a}$ and optionally part of the pronunciation decoder $\hat{\bm{\theta}}_{p}$. Note $\bar{\bm{p}}$ is never used as the training target as no new pronunciation knowledge is encoded within it. In a training batch, half of the training samples are sampled from $\mathcal{D}_{BS}$, while the other half are sampled from $\bar{\mathcal{D}}_{TS}$. The acoustic decoder is discarded after MTL.

Specifically, the acoustic decoder is integrated by attending to one of the intermediate representations (or \emph{levels}) of Seq2Seq frontend, including $\bm{e}_{1:L}$ (\emph{level-1)}, $\bm{a}_{1:T}$ (\emph{level-2}), $\bm{c}_{1:T}$ (\emph{level-3}), $\bm{d}_{1:T}$ (\emph{level-4}) or the sampled tokens' embeddings after applying Gumbel-Softmax \cite{Jang2017} to $\bm{l}_{1:T}$ (\emph{level-5}) (see Fig.\ \ref{fig:multitask_model}). We empirically find the best \emph{level} (i.e., the best multi-task model architecture) in Sec.\ \ref{experiments}. Note for \emph{levels 2-5}, $\bar{\bm{p}}_{0:T-1}$ is needed by the teacher-forcing inference of the pronunciation decoder to rapidly and robustly acquire $\bm{a}_{1:T}$, $\bm{c}_{1:T}$, $\bm{d}_{1:T}$ and $\bm{l}_{1:T}$ respectively, considering the auto-regressive inference is both slow and non-robust during training. In contrast, for \emph{level-1}, $\bar{\bm{p}}_{0:T-1}$ is not needed to acquire $\bm{e}_{1:L}$, which eliminates the need for the pre-training stage in Fig.\ \ref{fig:flowchart}.

\section{Experiments}
\label{experiments}


We empirically compare the performance of different multi-task model architectures plus also compare mel-spectrogram with MFCCs when used as target acoustic features within the method. We also compare performance to that of a baseline Seq2Seq frontend trained only with  $\mathcal{D}_{BS}$, and a Seq2Seq frontend trained with  $\mathcal{D}_{BS}$ and improved by the FA method with $\mathcal{D}_{TS}$ (re-implemented by us following \cite{Sun2023}). The experiments are based on standard British English (RPX).

\subsection{Experimental Setup}
\label{setup}

For $\mathcal{D}_{BS}$, we use the normalised text of three training subsets of LibriSpeech \cite{Panayotov2015} as the unlabeled text. We keep sentences that contain no out-of-dictionary words, amounting to 206k sentences. The text of Dev-clean forms the unlabelled text for validation (2.7k sentences). Festival \cite{Clark2007} is used as our pipeline-based frontend to create $\mathcal{D}_{BS}$. \texttt{unilex-rpx} \cite{Fitt2000} is used as the built-in dictionary and defines the pronunciation token set. We follow the pre-processing steps in \cite{Sun2023}.

For $\mathcal{D}_{TS}$, text-audio pairs of three RPX speakers of Hi-Fi TTS \cite{Bakhturina2021} (\#92, 6097 and 9136) are merged, amounting to 103k sentences (81.7 hours). We use vctk.hifigan.v1 \cite{Hayashi2023} to extract mel-spectrograms, whose sample rate/\# filter banks/window size/hop size are 24kHz/80/50ms/12.5ms. We use HTK \cite{Young2015} to extract MFCCs, whose sample rate/\# Cepstrum components/window size/hop size are 16kHz/13/25ms/10ms.

During evaluation, we focus on word-level performance after segmentation. Besides phone error rate (PER), we also report word accuracy considering phones only (WAccP) and word accuracy considering phones, stresses and syllable boundaries altogether (WAcc). All metrics are computed on word tokens.
Furthermore, we distinguish between \emph{main-covered} words (words covered in $\mathcal{D}_{BS}$), \emph{extra-exclusive} words (words covered in $\mathcal{D}_{TS}$ but not covered in $\mathcal{D}_{BS}$) and OOV words (words not covered in either set). As noted earlier, we mainly focus on evaluating the generalization on the \emph{main task} to \emph{extra-exclusive} words. To this end, we run $\forall \bm{x}^{(j)} \in \mathcal{D}_{TS}$ through Festival to get the ground-truth pronunciations to form our first test set, which includes 1.6k \emph{extra-exclusive} word tokens (11.6k phones). We also evaluate the memorization of \emph{main-covered} words on this test set.
Though the generalization to OOV words is not our main focus, we still evaluate it in this paper. To that end, we merge those Hi-Fi TTS sentences  that contain OOV words (and no out-of-dictionary words) and run them through Festival to form our second test set, which includes 4.4k sentences and equiv.\ 2.7k OOV word tokens (19.6k phones).

All the evaluated models share the same hyperparameters. The encoder and the two decoders each have 2 layers. The embedding dimension is 256. The hidden dimension is 512. The number of LMM mixture components is 2. The dropout rate is 0.3. The number of mixture components in GMM attention is 5. The learning rate is 5e-5. Adam is used as the optimizer. The batch size is 36. 
We pick $\lambda \in \{0.1,1.0\}$ based on \emph{extra-exclusive} word performance on the validation set. The beam size is set to 30. 

\subsection{Results}
\label{results}

\begin{table}[!t]
\caption{Extra-exclusive word token results of various MTL configurations.  `enc/attn/ctx/dec/logit' corresponds to the levels 1-5 in Fig.\ \ref{fig:multitask_model}, respectively. The number in bold indicates it is not significantly different from the best value (underscored) ($p>0.05$).}
\label{table:result}
\centering
\footnotesize
\setlength{\tabcolsep}{3pt}
\begin{tabular}{llp{35pt}ll}
\hline
Method / Model & Acou. feat.  & PER (\%) & WAccP (\%) & WAcc (\%) \\
\hline
Baseline & - & $2.5$ & $88.0$ & $82.3$ \\
FA method \cite{Sun2023} & MFCCs & $\underline{\mathbf{1.5}}$ & $\mathbf{92.5}$ & $\mathbf{85.9}$ \\
\hline
L1 enc & Mel & $\mathbf{1.6}$ & $\underline{\mathbf{92.6}}$ & $\mathbf{85.8}$ \\
L1 attn &  & $\mathbf{1.8}$ & $\mathbf{92.5}$ & $\underline{\mathbf{86.0}}$ \\
L1 ctx &  & $2.2$ & $88.5$ & $81.8$ \\
L1 dec &  & $\mathbf{1.8}$ & $\mathbf{92.1}$ & $\mathbf{85.6}$ \\
L1 logit &  & $3.6$ & $78.4$ & $72.3$ \\
L2 enc &  & $2.3$ & $88.5$ & $81.9$ \\
L2 attn &  & $2.0$ & $\mathbf{90.7}$ & $83.3$ \\
L2 ctx &  & $2.1$ & $90.4$ & $\mathbf{83.7}$ \\
L2 dec &  & $2.5$ & $86.5$ & $80.3$ \\
L2 logit &  & $3.4$ & $79.0$ & $72.1$ \\
LMM+NLL enc &  & $2.1$ & $89.8$ & $83.5$ \\
LMM+NLL ctx &  & $3.4$ & $81.0$ & $73.1$ \\
LMM+NLL logit &  & $3.3$ & $79.1$ & $72.6$ \\
\hline
L1 enc & MFCCs & $2.7$ & $85.2$ & $77.6$ \\
L1 attn &  & $2.1$ & $89.2$ & $81.8$ \\
L1 ctx &  & $3.3$ & $80.4$ & $73.1$ \\
L2 enc &  & $2.7$ & $86.2$ & $78.4$ \\
L2 ctx &  & $3.6$ & $78.8$ & $71.1$ \\
\hline
\end{tabular}
\end{table}

For \emph{main-covered} words, all models listed in Table \ref{table:result} achieve similarly very high performance ($\text{PER}<=0.03\%$, $\text{WAccP}>99.9\%$, $\text{WAcc}>99.9\%$), matching those in \cite{Sun2023}.

For \emph{extra-exclusive} words, the results are shown in Table \ref{table:result}. Both the FA method ($1.5\%/92.5\%/85.9\%$) and the MTL-based method ($1.6\%/92.6\%/86.0\%$) significantly outperform the baseline ($2.5\%/88\%/82.3\%$) on PER, WAccP and WAcc, respectively ($p<0.05$). The difference between the MTL-based method and the FA method is not significant. Among all MTL configurations, `L1 enc Mel' performs best on PER and WAccP, `L1 attn Mel' performs best on WAcc, and no significant difference exists between `L1 enc Mel', `L1 attn Mel' and `L1 dec Mel'.
When using mel-spectrograms as the acoustic features in the MTL-based method, we see for $\mathcal{L}_{a}$, overall L1 loss is significantly better than L2 loss. L1 loss is also significantly better than its multimodal counterpart LMM+NLL, indicating modelling multimodality does not help here. Finally, when used as the acoustic features in the MTL-based method, mel-spectrogram is significantly better than MFCCs.
The impressive generalization on the \emph{main task} to \emph{extra-exclusive} words is consistent with \cite{Sun2024} and suggests the MTL-based method can be used as an alternative method to the FA method to acquire pronunciation knowledge from transcribed speech audio to improve Seq2Seq frontends.

For \emph{OOV} words, the results are shown in Table \ref{table:result_oov}. Both the FA method and the MTL-based method slightly underperform the baseline, with the latter having a statistically significant difference in PER/WAccP.
MTL can underperform single-task learning (STL) in some cases, which has been widely shown \cite{Caruana1998, Wu2020, Kang2024}. In \cite{Kang2024}, an MTL model jointly modelling TN, Part-of-Speech (POS) tagging and homograph disambiguation underperforms a STL TN model and a STL POS tagger, respectively.
Even so, MTL-improved Seq2Seq frontends may still be used offline to predict the pronunciations for \emph{extra-exclusive} words and then expand the dictionary.
We will work on improving the generalization to OOV words in future work. 

\begin{table}[!t]
\caption{Out-of-vocabulary (OOV) word token results. The number in bold indicates it is not significantly different from Baseline}
\label{table:result_oov}
\centering
\footnotesize
\setlength{\tabcolsep}{3pt}
\begin{tabular}{lllll}
\hline
Model & Acou. feat.  & PER (\%) & WAccP (\%) & WAcc (\%) \\
\hline
Baseline & - & $\mathbf{2.9}$ & $\mathbf{87.1}$ & $\mathbf{82.1}$ \\
FA method \cite{Sun2023} & MFCCs & $\mathbf{2.9}$ & $\mathbf{86.1}$ & $\mathbf{81.3}$ \\
L1 enc & Mel & $3.5$ & $84.6$ & $\mathbf{80.6}$ \\
\hline
\end{tabular}
\end{table}

\subsection{Analyses}
\label{analyses}

We verify the improved generalization to \emph{extra-exclusive} words is indeed due to MTL transferring pronunciation knowledge encoded in the training targets of \emph{extra task} to the \emph{main task}, by ruling out two main alternative explanations \cite{Caruana1998}, which are (i) the effective network capacity being reduced by parameter sharing and (ii) the \emph{extra task} training targets acting as a source of noise.
For (i), we evaluate STL baselines at various reduced network sizes (hidden dim $\in \{128, 256, 384\}$), as shown in Table \ref{table:analyses} (middle). All three underperform the original baseline and our best MTL configuration `L1 enc Mel', showing the improvement is not due to reduced effective network capacity. For (ii), we shuffle the targets among all the cases in $\mathcal{D}_{TS}$, i.e., create a shuffled set $\{(\bm{x}^{(j)}, \bm{m}^{(k)}) \,|\, k \neq j\}_{j=1}^M$, breaking \emph{extra task}'s relatedness to the \emph{main task} ($\bm{m}^{(k)}$ now being a source of noise) while keeping \emph{extra task}'s target distribution unchanged. `L1 enc Mel' trained with shuffled $\mathcal{D}_{TS}$ (last part of Table \ref{table:analyses}) underperforms `L1 enc Mel' significantly and is comparable to the baseline, showing the original matched $\bm{m}^{(j)}$ is far more than just a source of noise.

\begin{table}[!t]
\caption{Analysis results on extra-exclusive word tokens.}
\label{table:analyses}
\centering
\footnotesize
\setlength{\tabcolsep}{3pt}
\begin{tabular}{lllll}
\hline
Model & Acou. feat.  & PER (\%) & WAccP (\%) & WAcc (\%) \\
\hline
L1 enc & Mel & $\mathbf{1.6}$ & $\mathbf{92.6}$ & $\mathbf{85.8}$ \\
Baseline & - & $2.5$ & $88.0$ & $82.3$ \\
\hline
Baseline, h=384 & - & $2.7$ & $87.0$ & $80.5$ \\
Baseline, h=256 & - & $2.8$ & $86.1$ & $79.2$ \\
Baseline, h=128 & - & $5.9$ & $67.2$ & $58.3$ \\
\hline
L1 enc, shuffled & Mel & $2.3$ & $88.7$ & $82.2$ \\
\hline
\end{tabular}
\end{table}

\section{Conclusion}
\label{conclusion}

We have proposed a novel MTL-based method for acquiring pronunciation knowledge from transcribed speech audio to improve Seq2Seq frontends. 
The proposed method only requires a slight adaptation of the current Seq2Seq frontend model, by integrating an auxiliary acoustic decoder, which is discarded after MTL. It avoids ASR training and decoding. Experiments show our method is very successful in transferring pronunciation knowledge encoded in speech audio to Seq2Seq frontends (reducing PER by 36\% relative for words covered exclusively in speech), achieving a similar performance to the previous method while having a much simpler implementation flow. 


\clearpage
\bibliographystyle{IEEEtran}
\bibliography{paper}

\begin{thebibliography}{10}
\providecommand{\url}[1]{#1}
\csname url@samestyle\endcsname
\providecommand{\newblock}{\relax}
\providecommand{\bibinfo}[2]{#2}
\providecommand{\BIBentrySTDinterwordspacing}{\spaceskip=0pt\relax}
\providecommand{\BIBentryALTinterwordstretchfactor}{4}
\providecommand{\BIBentryALTinterwordspacing}{\spaceskip=\fontdimen2\font plus
\BIBentryALTinterwordstretchfactor\fontdimen3\font minus
  \fontdimen4\font\relax}
\providecommand{\BIBforeignlanguage}[2]{{%
\expandafter\ifx\csname l@#1\endcsname\relax
\typeout{** WARNING: IEEEtran.bst: No hyphenation pattern has been}%
\typeout{** loaded for the language `#1'. Using the pattern for}%
\typeout{** the default language instead.}%
\else
\language=\csname l@#1\endcsname
\fi
#2}}
\providecommand{\BIBdecl}{\relax}
\BIBdecl

\bibitem{Fong2019}
\BIBentryALTinterwordspacing
J.~Fong, J.~Taylor, K.~Richmond, and S.~King, ``A comparison of letters and
  phones as input to sequence-to-sequence models for speech synthesis,'' in
  \emph{Proc. 10th ISCA Speech Synthesis Workshop}, 2019, pp. 223--227.
  [Online]. Available: \url{http://dx.doi.org/10.21437/SSW.2019-40}
\BIBentrySTDinterwordspacing

\bibitem{Shen2020}
J.~Shen, Y.~Jia, M.~Chrzanowski, Y.~Zhang, I.~Elias, H.~Zen, and Y.~Wu,
  ``Non-attentive {T}acotron: Robust and controllable neural {TTS} synthesis
  including unsupervised duration modeling,'' \emph{ArXiv}, vol.
  abs/2010.04301, 2020.

\bibitem{Ren2021}
\BIBentryALTinterwordspacing
Y.~Ren, C.~Hu, X.~Tan, T.~Qin, S.~Zhao, Z.~Zhao, and T.-Y. Liu, ``{FastSpeech}
  2: Fast and high-quality end-to-end text to speech,'' in \emph{International
  Conference on Learning Representations}, 2021. [Online]. Available:
  \url{https://openreview.net/forum?id=piLPYqxtWuA}
\BIBentrySTDinterwordspacing

\bibitem{Tan2021}
X.~Tan, T.~Qin, F.~K. Soong, and T.-Y. Liu, ``A survey on neural speech
  synthesis,'' \emph{ArXiv}, vol. abs/2106.15561, 2021.

\bibitem{Fitt2000}
\BIBentryALTinterwordspacing
S.~Fitt, ``Documentation and user guide to {UNISYN} lexicon and post-lexical
  rules,'' 2000. [Online]. Available:
  \url{https://www.cstr.ed.ac.uk/projects/unisyn/}
\BIBentrySTDinterwordspacing

\bibitem{Conkie2020}
A.~Conkie and A.~M. Finch, ``Scalable multilingual frontend for {TTS},''
  \emph{ICASSP 2020 - 2020 IEEE International Conference on Acoustics, Speech
  and Signal Processing (ICASSP)}, pp. 6684--6688, 2020.

\bibitem{Pan2020}
J.~Pan, X.~Yin, Z.~Zhang, S.~Liu, Y.~Zhang, Z.~Ma, and Y.~Wang, ``A unified
  sequence-to-sequence front-end model for {M}andarin text-to-speech
  synthesis,'' in \emph{ICASSP 2020 - 2020 IEEE International Conference on
  Acoustics, Speech and Signal Processing (ICASSP)}, 2020, pp. 6689--6693.

\bibitem{Sun2023}
S.~Sun, K.~Richmond, and H.~Tang, ``Improving {Seq2Seq} {TTS} frontends with
  transcribed speech audio,'' \emph{IEEE/ACM Transactions on Audio, Speech, and
  Language Processing}, vol.~31, pp. 1940--1952, 2023.

\bibitem{Comini2023}
G.~Comini, S.~Ribeiro, F.~Yang, H.~Shim, and J.~Lorenzo-Trueba, ``Multilingual
  context-based pronunciation learning for text-to-speech,'' in \emph{Proc.
  INTERSPEECH 2023}, 2023, pp. 631--635.

\bibitem{Caruana1998}
\BIBentryALTinterwordspacing
R.~Caruana, \emph{Multitask Learning}.\hskip 1em plus 0.5em minus 0.4em\relax
  Boston, MA: Springer US, 1998, pp. 95--133. [Online]. Available:
  \url{https://doi.org/10.1007/978-1-4615-5529-2_5}
\BIBentrySTDinterwordspacing

\bibitem{Sun2024}
S.~Sun and K.~Richmond, ``Learning pronunciation from other accents via
  pronunciation knowledge transfer,'' in \emph{Interspeech 2024}, 2024, pp.
  2805--2809.

\bibitem{Rezackova2021}
M.~Řezáčková, J.~Švec, and D.~Tihelka, ``{T5G2P}: Using text-to-text
  transfer transformer for grapheme-to-phoneme conversion,'' in \emph{Proc.
  Interspeech 2021}, 2021, pp. 6--10.

\bibitem{Ploujnikov2022}
A.~Ploujnikov and M.~Ravanelli, ``{SoundChoice}: Grapheme-to-phoneme models
  with semantic disambiguation,'' in \emph{Proc. Interspeech 2022}, 2022, pp.
  486--490.

\bibitem{Battenberg2020}
E.~Battenberg, R.~Skerry-Ryan, S.~Mariooryad, D.~Stanton, D.~Kao, M.~Shannon,
  and T.~Bagby, ``Location-relative attention mechanisms for robust long-form
  speech synthesis,'' in \emph{ICASSP 2020 - 2020 IEEE International Conference
  on Acoustics, Speech and Signal Processing (ICASSP)}, 2020, pp. 6194--6198.

\bibitem{Rutherford2014}
A.~T. Rutherford, F.~Peng, and F.~Beaufays, ``Pronunciation learning for
  named-entities through crowd-sourcing,'' in \emph{Proc. Interspeech 2014},
  2014, pp. 1448--1452.

\bibitem{Kou2015}
Z.~Kou, D.~Stanton, F.~Peng, F.~Beaufays, and T.~Strohman, ``Fix it where it
  fails: Pronunciation learning by mining error corrections from speech logs,''
  in \emph{2015 IEEE International Conference on Acoustics, Speech and Signal
  Processing (ICASSP)}, 2015, pp. 4619--4623.

\bibitem{Bruguier2016}
A.~Bruguier, F.~Peng, and F.~Beaufays, ``Learning personalized pronunciations
  for contact name recognition,'' in \emph{Proc. Interspeech 2016}, 2016, pp.
  3096--3100.

\bibitem{Bruguier2017}
A.~Bruguier, D.~Gnanapragasam, L.~Johnson, K.~Rao, and F.~Beaufays,
  ``Pronunciation learning with {RNN}-{T}ransducers,'' in \emph{Proc.
  Interspeech 2017}, 2017, pp. 2556--2560.

\bibitem{Ribeiro2023}
S.~Ribeiro, G.~Comini, and J.~Lorenzo-Trueba, ``Improving grapheme-to-phoneme
  conversion by learning pronunciations from speech recordings,'' in
  \emph{Proc. INTERSPEECH 2023}, 2023, pp. 999--1003.

\bibitem{Milde2017}
B.~Milde, C.~Schmidt, and J.~Köhler, ``Multitask sequence-to-sequence models
  for grapheme-to-phoneme conversion,'' in \emph{Proc. Interspeech 2017}, 2017,
  pp. 2536--2540.

\bibitem{Route2019}
\BIBentryALTinterwordspacing
J.~Route, S.~Hillis, I.~Czeresnia~Etinger, H.~Zhang, and A.~W. Black,
  ``Multimodal, multilingual grapheme-to-phoneme conversion for low-resource
  languages,'' in \emph{Proceedings of the 2nd Workshop on Deep Learning
  Approaches for Low-Resource NLP (DeepLo 2019)}.\hskip 1em plus 0.5em minus
  0.4em\relax Hong Kong, China: Association for Computational Linguistics, Nov.
  2019, pp. 192--201. [Online]. Available:
  \url{https://aclanthology.org/D19-6121}
\BIBentrySTDinterwordspacing

\bibitem{Thompson2022}
\BIBentryALTinterwordspacing
B.~Thompson and A.~Alshehri, ``Improving {A}rabic diacritization by learning to
  diacritize and translate,'' in \emph{Proceedings of the 19th International
  Conference on Spoken Language Translation (IWSLT 2022)}, E.~Salesky,
  M.~Federico, and M.~Costa-juss{\`a}, Eds.\hskip 1em plus 0.5em minus
  0.4em\relax Dublin, Ireland (in-person and online): Association for
  Computational Linguistics, May 2022, pp. 11--21. [Online]. Available:
  \url{https://aclanthology.org/2022.iwslt-1.2}
\BIBentrySTDinterwordspacing

\bibitem{Ying2024}
Z.~Ying, C.~Li, Y.~Dong, Q.~Kong, Q.~Tian, Y.~Huo, and Y.~Wang, ``A unified
  front-end framework for {English} text-to-speech synthesis,'' in \emph{ICASSP
  2024 - 2024 IEEE International Conference on Acoustics, Speech and Signal
  Processing (ICASSP)}, 2024, pp. 10\,181--10\,185.

\bibitem{Kang2024}
W.~Kang, Y.~Wang, S.~Zhang, A.~Hinsvark, and Q.~He, ``Multi-task learning for
  front-end text processing in {TTS},'' in \emph{ICASSP 2024 - 2024 IEEE
  International Conference on Acoustics, Speech and Signal Processing
  (ICASSP)}, 2024, pp. 10\,796--10\,800.

\bibitem{Peters2017}
\BIBentryALTinterwordspacing
B.~Peters, J.~Dehdari, and J.~van Genabith, ``Massively multilingual neural
  grapheme-to-phoneme conversion,'' in \emph{Proceedings of the First Workshop
  on Building Linguistically Generalizable {NLP} Systems}.\hskip 1em plus 0.5em
  minus 0.4em\relax Copenhagen, Denmark: Association for Computational
  Linguistics, Sep. 2017, pp. 19--26. [Online]. Available:
  \url{https://aclanthology.org/W17-5403}
\BIBentrySTDinterwordspacing

\bibitem{Kim2022}
\BIBentryALTinterwordspacing
H.-Y. Kim, J.-H. Kim, and J.-M. Kim, ``Fast bilingual grapheme-to-phoneme
  conversion,'' in \emph{Proceedings of the 2022 Conference of the North
  American Chapter of the Association for Computational Linguistics: Human
  Language Technologies: Industry Track}.\hskip 1em plus 0.5em minus
  0.4em\relax Hybrid: Seattle, Washington + Online: Association for
  Computational Linguistics, Jul. 2022, pp. 289--296. [Online]. Available:
  \url{https://aclanthology.org/2022.naacl-industry.32}
\BIBentrySTDinterwordspacing

\bibitem{Zhu2022}
J.~Zhu, C.~Zhang, and D.~Jurgens, ``{ByT5} model for massively multilingual
  grapheme-to-phoneme conversion,'' in \emph{Proc. Interspeech 2022}, 2022, pp.
  446--450.

\bibitem{Shen2018}
J.~Shen, R.~Pang, R.~J. Weiss, M.~Schuster, N.~Jaitly, Z.~Yang, Z.~Chen,
  Y.~Zhang, Y.~Wang, R.~Skerry-Ryan, R.~A. Saurous, Y.~Agiomyrgiannakis, and
  Y.~Wu, ``Natural {TTS} synthesis by conditioning {Wavenet} on {MEL}
  spectrogram predictions,'' in \emph{2018 IEEE International Conference on
  Acoustics, Speech and Signal Processing (ICASSP)}, 2018, pp. 4779--4783.

\bibitem{Ren2022}
\BIBentryALTinterwordspacing
Y.~Ren, X.~Tan, T.~Qin, Z.~Zhao, and T.-Y. Liu, ``Revisiting over-smoothness in
  text to speech,'' in \emph{Proceedings of the 60th Annual Meeting of the
  Association for Computational Linguistics (Volume 1: Long Papers)},
  S.~Muresan, P.~Nakov, and A.~Villavicencio, Eds.\hskip 1em plus 0.5em minus
  0.4em\relax Dublin, Ireland: Association for Computational Linguistics, May
  2022, pp. 8197--8213. [Online]. Available:
  \url{https://aclanthology.org/2022.acl-long.564}
\BIBentrySTDinterwordspacing

\bibitem{Kogel2023}
F.~Kögel, B.~Nguyen, and F.~Cardinaux, ``Towards robust {FastSpeech} 2 by
  modelling residual multimodality,'' in \emph{Proc. INTERSPEECH 2023}, 2023,
  pp. 4309--4313.

\bibitem{Jang2017}
\BIBentryALTinterwordspacing
E.~Jang, S.~Gu, and B.~Poole, ``Categorical reparameterization with
  {Gumbel-Softmax},'' in \emph{International Conference on Learning
  Representations}, 2017. [Online]. Available:
  \url{https://openreview.net/forum?id=rkE3y85ee}
\BIBentrySTDinterwordspacing

\bibitem{Panayotov2015}
V.~Panayotov, G.~Chen, D.~Povey, and S.~Khudanpur, ``{LibriSpeech}: An {ASR}
  corpus based on public domain audio books,'' in \emph{2015 IEEE International
  Conference on Acoustics, Speech and Signal Processing (ICASSP)}, 2015, pp.
  5206--5210.

\bibitem{Clark2007}
\BIBentryALTinterwordspacing
R.~A. Clark, K.~Richmond, and S.~King, ``Multisyn: Open-domain unit selection
  for the {F}estival speech synthesis system,'' \emph{Speech Communication},
  vol.~49, no.~4, pp. 317 -- 330, 2007. [Online]. Available:
  \url{http://www.sciencedirect.com/science/article/pii/S0167639307000398}
\BIBentrySTDinterwordspacing

\bibitem{Bakhturina2021}
E.~Bakhturina, V.~Lavrukhin, B.~Ginsburg, and Y.~Zhang, ``{Hi-Fi} multi-speaker
  {E}nglish {TTS} dataset,'' in \emph{Proc. Interspeech 2021}, 2021, pp.
  2776--2780.

\bibitem{Hayashi2023}
T.~Hayashi, ``Parallel wavegan implementation with pytorch,''
  \url{https://github.com/kan-bayashi/ParallelWaveGAN}, 2023.

\bibitem{Young2015}
\BIBentryALTinterwordspacing
S.~Young, G.~Evermann, M.~Gales, T.~Hain, D.~Kershaw, X.~Liu, G.~Moore,
  J.~Odell, D.~Ollason, D.~Povey, A.~Ragni, V.~Valtchev, P.~Woodland, and
  C.~Zhang, \emph{The HTK Book (version 3.5a)}, 2015. [Online]. Available:
  \url{https://htk.eng.cam.ac.uk/docs/docs.shtml}
\BIBentrySTDinterwordspacing

\bibitem{Wu2020}
\BIBentryALTinterwordspacing
S.~Wu, H.~R. Zhang, and C.~Ré, ``Understanding and improving information
  transfer in multi-task learning,'' in \emph{International Conference on
  Learning Representations}, 2020. [Online]. Available:
  \url{https://openreview.net/forum?id=SylzhkBtDB}
\BIBentrySTDinterwordspacing

\end{thebibliography}

\end{document}